\documentclass[letterpaper, 10 pt, conference]{IEEEtran} 

\IEEEoverridecommandlockouts                              

\usepackage{amsmath,amsfonts}
\usepackage{algorithmic}
\usepackage{algorithm}
\usepackage{array}
\usepackage{textcomp}
\usepackage{stfloats}
\usepackage{url}
\usepackage{verbatim}
\usepackage{graphicx}
\usepackage{cite}
\usepackage{enumitem} 

\usepackage{array}
\usepackage{colortbl}
\usepackage[dvipsnames]{xcolor}
\usepackage{makecell}
\usepackage{tabu,multirow}
\newcommand{\refFig}[1]{Fig.~\ref{#1}}
\newcommand{\refTab}[1]{TABLE~\ref{#1}}

\usepackage{colortbl} 
\usepackage{xcolor}   
\usepackage{tabu}

\title{LangPert: Detecting and Handling Task-level Perturbations for Robust Object Rearrangement
}

\author{Xu Yin\textsuperscript{1}, Min-Sung Yoon\textsuperscript{1}, Yuchi Huo\textsuperscript{2}, Kang Zhang\textsuperscript{3}, and Sung-Eui Yoon\textsuperscript{\textdagger}%
\thanks{
\textsuperscript{1}Xu Yin and Min-Sung Yoon are with the School of Computing, Korea Advanced Institute of Science and Technology (KAIST), Daejeon 34141, South Korea ({\tt\footnotesize \{yinofsgvr, minsung.yoon\}@kaist.ac.kr}).}%
\thanks{\textsuperscript{2}Yuchi Huo is with the State Key Lab of CAD and CG, Zhejiang University, China and Zhejiang Lab, China 310058 ({\tt\footnotesize huo.yuchi.sc@gmail.com}).}%
\thanks{\textsuperscript{3}Kang Zhang is with the School of Electrical Engineering, KAIST. ({\tt\footnotesize zhangkang@kaist.ac.kr}).}%
\thanks{{\textsuperscript{\textdagger}}Sung-Eui Yoon (corresponding author) is with the Faculty of School of Computing, Korea Advanced Institute of Science and Technology, Deajeon 34141, South Korea ({\tt\footnotesize sungeui@gmail.com}).}%
}

\begin{document}
    \maketitle
    \thispagestyle{empty}
    \pagestyle{empty}
    
    \begin{abstract}
    Task execution for object rearrangement could be challenged by Task-Level Perturbations (TLP)—unexpected object additions, removals, and displacements—that can disrupt underlying visual policies and fundamentally compromise task feasibility and progress. To address these challenges, we present LangPert, a language-based framework designed to detect and mitigate TLP situations in tabletop rearrangement tasks. LangPert integrates a Visual Language Model (VLM) to comprehensively monitor policy's skill execution and environmental TLP, while leveraging the Hierarchical Chain-of-Thought (HCoT) reasoning mechanism to enhance the Large Language Model (LLM)'s contextual understanding and generate adaptive, corrective skill-execution plans. Our experimental results demonstrate that LangPert handles diverse TLP situations more effectively than baseline methods, achieving higher task completion rates, improved execution efficiency, and potential generalization to unseen scenarios.
 
    \end{abstract}
    \section{Introduction}
    \label{sec:introduction}
    Large Language Models (LLMs) have demonstrated significant potential as task planners~\cite{ground_decoding,saycan,inner}, effectively converting a user's task description into robot skill instructions. 
When combined with visual perception capabilities~\cite{GPT_4V} and trained on robot-specific datasets~\cite{rt2,sc_mllm}, LLMs can serve as both task planners and environment perceivers.
This enables robots to detect execution failures and re-plan skills in real-time~\cite{manipulate_anything, Corl_24, precondition3}, making LLMs a promising approach to enhance robotic autonomy in unstructured environments~\cite{reflect}.

However, most approaches focus on task completion in static environments, overlooking the dynamic nature of real-world scenarios where unexpected workspace changes could occur due to factors such as human intervention~\cite{palm}. These unpredictable disturbances~\cite{doremi} can significantly impact task progress and ultimately compromise task completability. Therefore, we need a robust intelligent system capable of detecting abnormal situations~\cite{aha} resulting from environmental changes, systematically reasoning about their effects on the overall task, and generating reliable recovery strategies to maintain execution continuity in the face of such perturbations.
 
\begin{figure}[t]
    \centering
    
    \includegraphics[width=\linewidth]{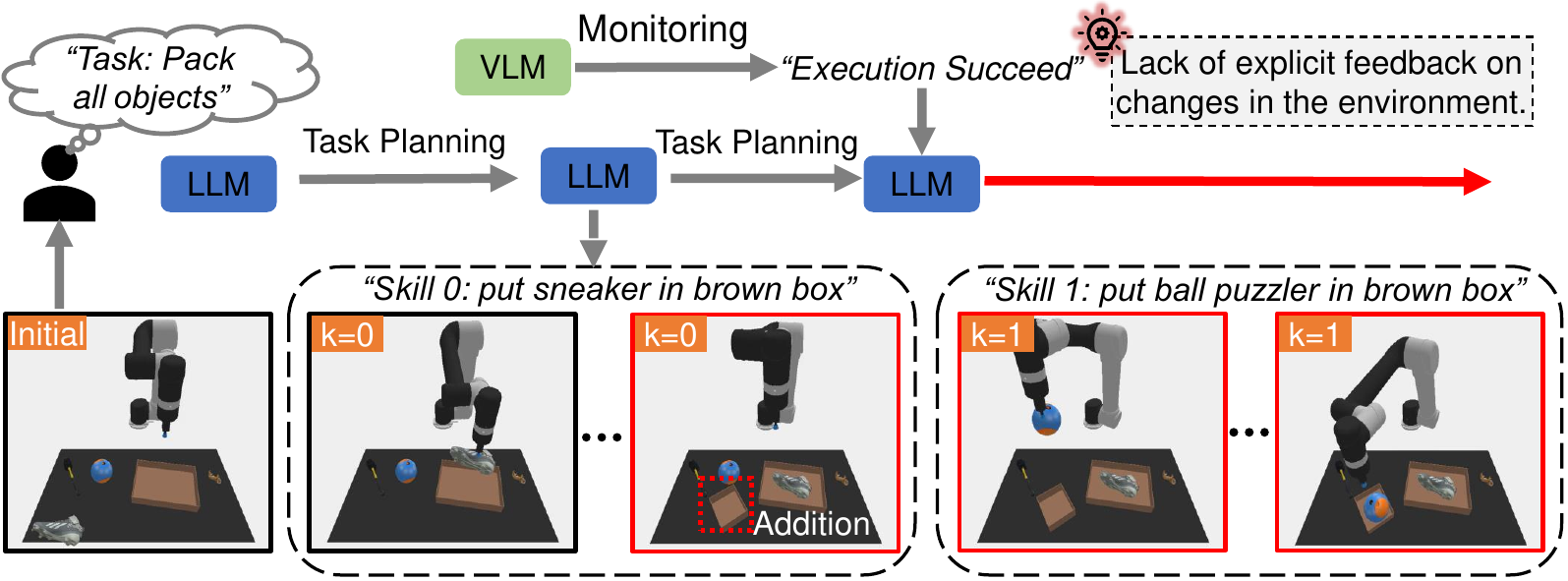}
    \vspace{-20pt}
    \caption{Environmental perturbation example in a box-packing task. As the robot places sneaker in brown box at step 0 ($k$=0), a new box unexpectedly appears (marked with the red dashed box). Standard VLM approaches~\cite{success_detector,doremi} detect only the robot’s execution outcome and lack awareness of broader environmental changes, leading the robot to place subsequent objects in the incorrect box. Handling such perturbations requires continuous global monitoring to detect perturbations and generate corrective strategies accordingly.
}
    \label{fig:figure_1}
    \vspace{-15pt}
\end{figure}

Among various challenges in dynamic environments, we focus on Task-Level Perturbations (TLP)—unexpected additions, removals, or displacements of objects—that disrupt immediate skill execution and overall task completion, particularly in tabletop rearrangement tasks. For example, as shown in Fig.~\ref{fig:figure_1}, the Visual Language Model (VLM) fails to discern a newly inserted box from the originally intended box that was present when the user issued the task description, causing the robot to misplace objects. This highlights the need for workspace awareness, reasoning capability, and adaptive re-planning to ensure robust task execution in dynamic environments.

Handling TLP is crucial for automated assembly systems, which frequently involve rearrangement operations to position components. Such disruptions can significantly impact workflows, leading to production delays, reduced efficiency, and even system breakdowns. These TLP often coincide with robotic execution failures, which can propagate errors across the production line and compromise system stability.

Addressing these challenges requires two key capabilities. First, a powerful perception model is needed to help the robot accurately distinguish between intended task state changes from robotic execution and unintended TLP. Second, the task planner must systematically assess these TLP impacts and generate corrective strategies to maintain task continuity.

In this work, we propose \textbf{LangPert}, a \textbf{Lang}uage-based framework to address task-level environmental \textbf{pert}urbations and execution failures in object rearrangement tasks. First, we categorize TLP into three types: addition, removal, and displacement. Each type affects the affordance prediction of visual policies, compromises overall task feasibility, or disrupts task progress. LangPert leverages a fine-tuned VLM as a global visual monitor, enabling precise detection of execution failure status and detailed descriptions of perturbations through language-based feedback. Besides, we design a prompt template that employs hierarchical chain-of-thought (HCoT) reasoning to comprehensively evaluate the impact of perturbations on task execution, facilitating reliable re-planning and generating corrective plans for informed decision-making

In experiments, our method effectively manages diverse perturbations and execution failures across tasks, outperforming existing baselines. The proposed HCoT reasoning enhances the LLM's contextual understanding, resulting in higher success rates and increased efficiency, particularly in complex or previously unseen scenarios.

    \section{Related Work}
    \label{sec:related_work}
    \noindent\textbf{LLM-based Task and Motion Planning.} Applying LLMs for task and motion planning has transformed the field of robotics, showing success across domains~\cite{rt1,rt2}. 
While various techniques~\cite{codepolicy,rt2,ground_decoding} address sophisticated tasks and enable dexterous manipulation, they often assume flawless execution or accessible human feedback for real-time adjustments~\cite{distilling,sc_mllm}, limiting their use in dynamic environments. 
In these environments, unexpected workspace changes (called TLP in this work)—including additions, removals, or displacements of objects—challenge task completion robustness. 
Our work both investigates the effects of these task-level perturbations in tabletop rearrangement tasks~\cite{cliport} and addresses the challenge by strengthening the interaction between the VLM monitor and the LLM planner, enabling the system to detect unintended changes and respond adaptively with corrective re-plans.

\noindent\textbf{Execution Failure \& Environment Perturbation.} Recently, LLM-based planners have shown strong self-correction capabilities~\cite{sc_mllm} by generating adjusted strategies to address failures in skill execution. For example, Reflect~\cite{reflect} uses multi-modal observations to infer failure causes, prompting the LLM for corrective re-planning. Doremi~\cite{doremi} pre-generates each planned instruction's visual constraints and conducts periodic verifications to ensure task completion. However, previous works focus on verifying and correcting isolated skill execution. In contrast, task-level perturbations (TLP)—independent of skill execution, caused by unexpected workspace changes, and capable of disrupting tasks—remain underexplored. Thus, we categorize and handle various TLP situations along with skill execution failure cases in tabletop rearrangement tasks.

\noindent\textbf{Dual-Model Approach in Dynamic Environments.} Existing LLM-based robotic planning frameworks~\cite{rt2} fall into two main categories: multimodal language models (MLM) to unify perception and planning~\cite{sc_mllm, manipulate_anything,palm,Corl_24}, and separate language models (dual models) for perceiving execution failure status~\cite{doremi} and recovery planning~\cite{precondition3}, connected via textual prompts. 
While the unified approach offers end-to-end integration, it requires extensive data for fine-tuning and faces challenges in grounding consecutive observations~\cite{video_survey}, limiting its effectiveness for tasks requiring close, long-term environmental tracking.
Therefore, we adopt the dual-model approach, assigning specific roles to each model: the VLM detects environmental anomalies and generates detailed perturbation descriptions by leveraging multimodal inputs, while the LLM-based planner generates adaptive, corrective plans through reasoning about the current situation. 
Our work also enhances perception through dual-view monitoring and improves planning via structured prompts with hierarchical reasoning, effectively handling diverse TLP and skill execution failure situations in dynamic environments (shown in Sec. \ref{sec:evaluation}).



    \section{Problem setup}
    \label{sec:task}
    \begin{figure}
    \centering
    \includegraphics[width=\linewidth]{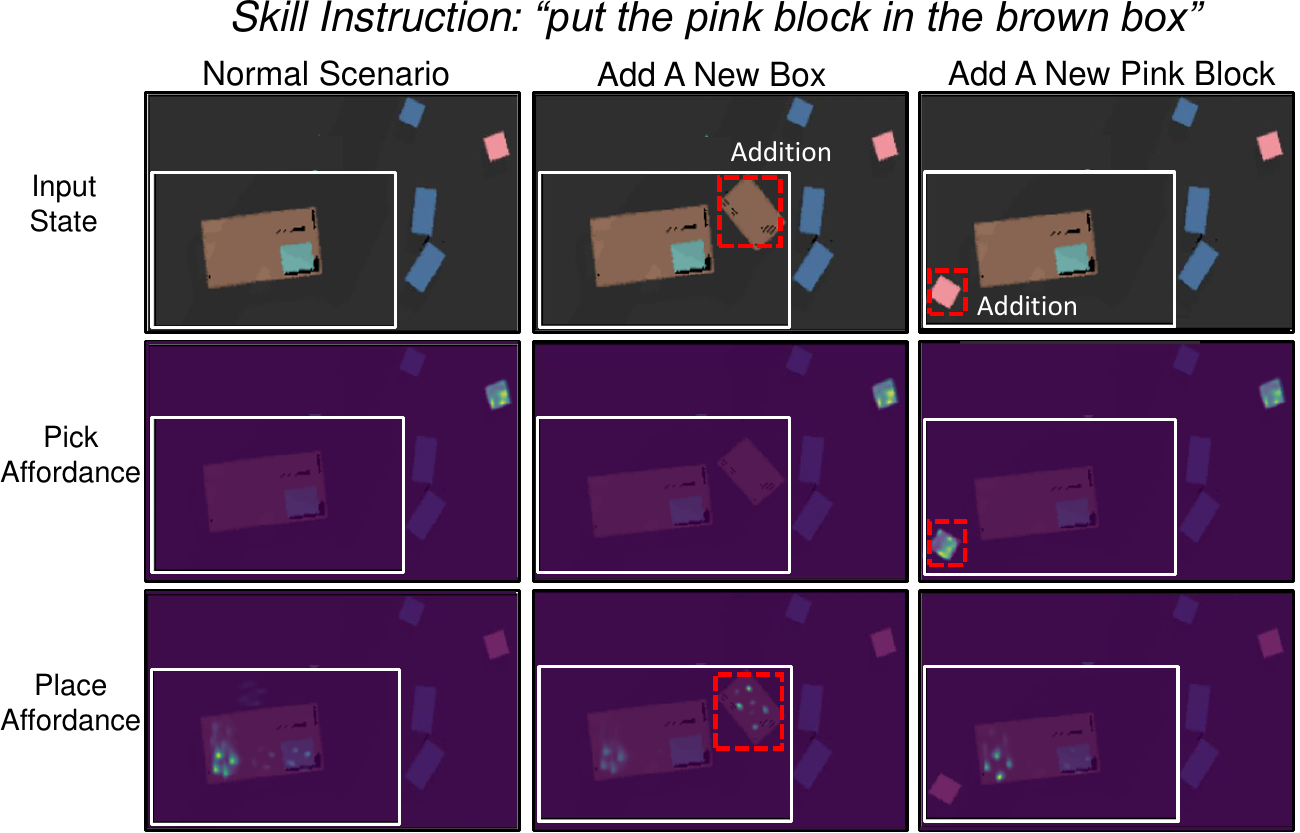}
    \vspace{-15pt}
    \caption{Illustration of how ADD perturbations affect affordance predictions. In the Normal Scenario (left), the original affordance predictions for placing pink block in brown box are shown, with pixel values indicating the probability of action success. In the Add A New Box and Add A New Pink Block scenarios (middle and right), the added objects (highlighted in red dashed boxes) alter these affordance distributions, potentially introducing ambiguity in execution.}
    \label{fig:perturbation_display}
    \vspace{-10pt}
\end{figure}
We investigate tabletop rearrangement tasks where robots manipulate objects into target configurations according to user-specified task descriptions. Additionally, we consider Task-Level Perturbations (TLP)—including unexpected object additions, removals, and displacements—that can disrupt task progress, create perceptual confusion, and even make task completion impossible.
The following description formalizes problem setups and categorizes TLP considered in this work.

\noindent{\textbf{Problem Formulation.}} We define these rearrangement tasks within a hierarchical framework that incorporates two levels of abstraction, allowing us to effectively manage the complexities inherent in task and motion planning for rearrangement tasks.

At the high level, given the task description $\mathcal{D}$ (e.g., “Put currently-seen blocks into brown box”), the LLM-based high-level task planner $\mathcal{T}$ generates skill instructions $\ell_k \in \mathcal{L}$ at each planning step $k$ (e.g., “Put red block into brown box”), where $\mathcal{L}$ is the skill repertoire executable by a low-level policy $\pi$.

At the low level, we use CLIPort~\cite{cliport} as the low-level policy $\pi$ to manipulate objects based on each skill instruction $\ell_k$. 
CLIPort processes visual observations $v_k \in \mathcal{V}$ alongside the corresponding instruction $\ell_k$ and generates the corresponding affordance maps for each skill, denoted as $\pi(v_k, \ell_k)$.
For instance, as shown in Fig.~\ref{fig:perturbation_display}, these affordance maps identify feasible locations for executing pick-and-place operations.


We define Task-Level Perturbations (TLP) as $\mathbf{E}$, representing workspace changes that occur independently of the skill execution, where each instance $e_k$ denotes a change occurring during the execution of skill instruction $\ell_k$ by the low-level policy $\pi$ at step $k$.
We categorize TLP $\mathbf{E}$ into three distinct types, each impacting tasks in a different way:

\begin{itemize}[leftmargin=0.33cm]
    \item \textbf{Addition (ADD):} The introduction of new objects, especially those visually similar to original objects, can alter the \textbf{\textit{affordance}} predictions of the low-level policy $\pi$, potentially leading the robot to interact with initially unintended objects. 
    To mitigate this, the high-level planner $\mathcal{T}$ need to assess whether the added objects interfere with task completion and generates corrective skill instruction plans, either discarding disruptive additions or ignoring non-obstructive ones to ensure uninterrupted, efficient task flow.
    
    \item \textbf{Removal (RMV):} The disappearance of objects can affect task \textbf{\textit{feasibility}}, especially when the missing object is essential for task completion. In such cases, the high-level planner $\mathcal{T}$ must evaluate its criticality, issuing an ``alert" to the user if the task becomes infeasible or continuing the task if the removal does not affect task completability.
    
    \item\textbf{Displacement (DIS):} Adversarial object relocation can disrupt task \textbf{\textit{progress}}, potentially misaligning the workspace state with the intended execution plan.
    To counteract such case, the high-level planner $\mathcal{T}$ needs to adaptively generate a new sequence of skill instructions to recover task progress.
\end{itemize}

Given that TLP $e_{k}$ and failures in low-level skill execution (e.g., dropping objects in the middle of manipulation) may occur at any time, a robust perception model is crucial for continuous workspace monitoring. It must detect unintended scene changes caused by $e_{k}$ and skill execution failures.
Furthermore, as each type of $E$ uniquely impacts task performance, the planner $\mathcal{T}$ must analyze these effects comprehensively and create adaptive, corrective strategies to ensure task continuity.


To further analyze the planner $\mathcal{T}$'s contextual understanding of different perturbations, we classify ADD and RMV into \textbf{task-related} and \textbf{distractor} types based on their relevance to the task description $\mathcal{D}$. 
The distractor types involve objects irrelevant to $\mathcal{D}$ (e.g., adding/removing a non-green block when $\mathcal{D}$ specifies “Pack all green blocks”), expecting $\mathcal{T}$ ignores this perturbation. 
In contrast, the task-related types involve objects essential to be handled (e.g., adding or removing green blocks), requiring $\mathcal{T}$ to either generate appropriate, corrective plans to discard added objects or issue an ``alert" signal. 
The high-level task planner $\mathcal{T}$ with better contextual understanding minimizes unnecessary manipulations of distractor objects, improving task efficiency and overall robustness.

    \section{Method}
    \label{sec:method}
    
\begin{figure}
    \centering
    \includegraphics[width=1\linewidth]{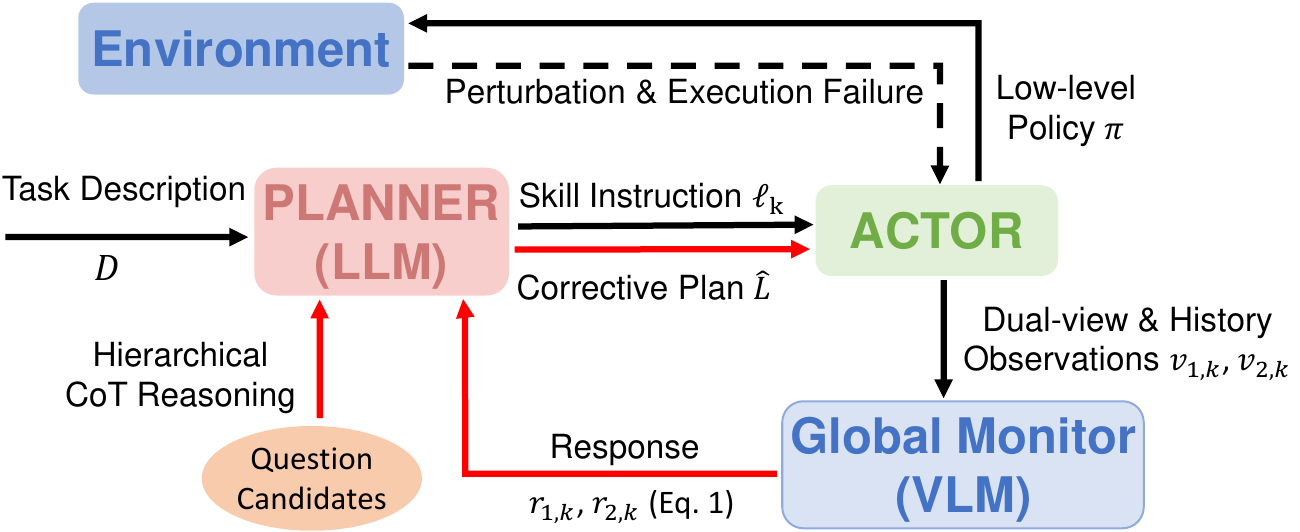}
    \vspace{-20pt}
    \caption{Framework overview. LangPert comprises: (1) a global VLM monitor that provides real-time workspace observations to detect execution failures and perturbations; (2) an LLM-based planner, which utilizes Hierarchical CoT reasoning to generate corrective plans based on VLM feedback; and (3) a language-conditioned actor module~\cite{cliport}, which executes low-level visual policies based on the skill instruction.}
    \label{fig:overview}
    \vspace{-8pt}
\end{figure}

\begin{figure}
    \centering
    \includegraphics[width=1\linewidth]{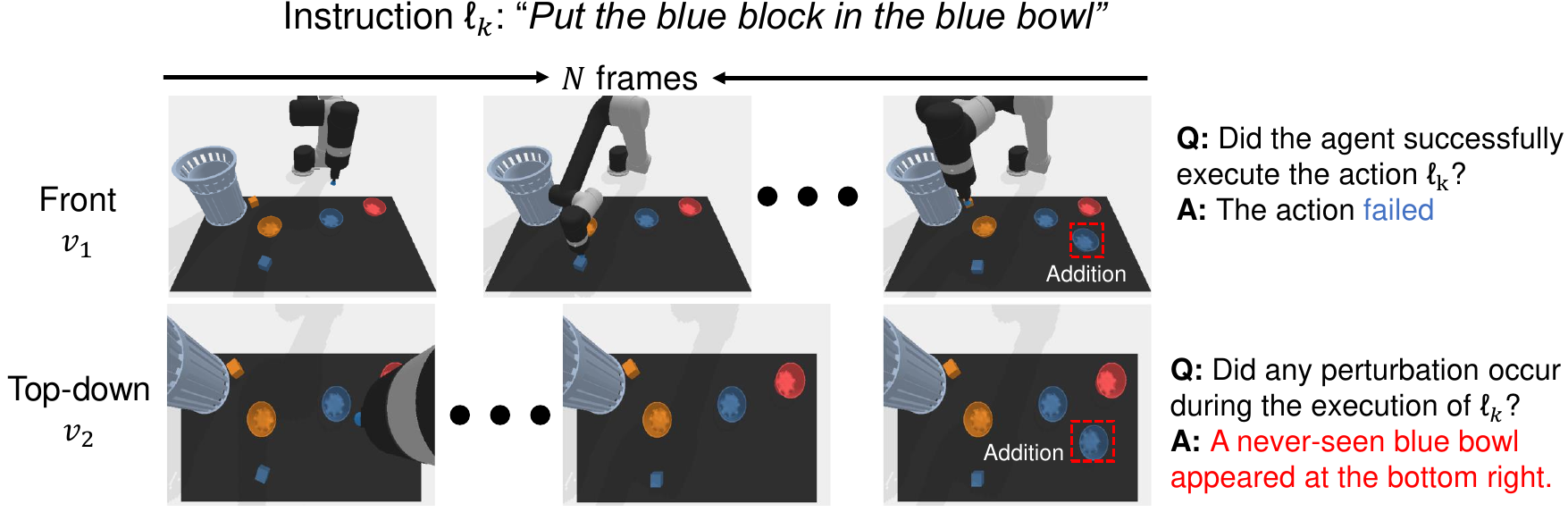}
    \vspace{-20pt}
    \caption{Camera configuration and VQA template for VLM. At each step $k$, given the skill instruction $\ell_{k}$, we capture RGB observations from both front and top-down views, forming two sequences of $N$ frames. The VLM is then queried to reason about the skill execution outcome and the perturbation.}
    \label{fig:vlm}
     \vspace{-13pt}
\end{figure}

We introduce \textbf{LangPert}, a \textbf{Lang}uage-based framework designed to address Task-Level \textbf{Pert}urbations (TLP) and skill execution failures during tabletop rearrangement tasks.
With advanced contextual understanding, LangPert detects unexpected changes, analyzes their impact, and produce adaptive, corrective plans, ensuring robust and efficient task completion.

\noindent\textbf{Framework Overview.} As illustrated in Fig.~\ref{fig:overview}, LangPert consists of three components: (1) a Vision Language Model (VLM) for real-time monitoring, (2) a LLM-based high-level task planner $\mathcal{T}$ with Hierarchical Chain-of-Thought (HCoT) reasoning for assessing unintended scene changes and generating adaptive re-plans, and (3) an actor, which is CLIPort~\cite{cliport} as the low-level policy $\pi$. 
At each planning step $k$, the VLM observes the workspace and detect the execution status of skill instructions $\ell_{k}$ and identifying any TLP $e_k$ or skill failures. 
Based on this detected information from the VLM, the planner $\mathcal{T}$ analyzes the impact of these disruptions and formulates adaptive corrective strategies to ensure seamless task progression and robustness in dynamic environments.

\subsection{VLM for Global Task Monitoring} 
\label{sec:vlm} 
Since the successful execution of each skill instruction $\ell_k$ by the actor $\pi$ results in intended workspace changes, it is critical to differentiate these from unintended perturbations $e_k$ and skill execution failures.
To address this challenge, we formulate both TLP detection and skill execution assessment as a Visual Question Answering (VQA) task, employing a fine-tuned VLM~\cite{openflamingo, vila, blip_3} as a global monitor. 
At each step $k$, the VLM evaluates the outcome of the skill instruction $\ell_k$ (i.e., success or failure) and generates structured descriptions of any detected task-level perturbations $e_k$, specifying both \textit{what} changed and \textit{where} it occurred.

\noindent\textbf{Dual-View Perception.} To ensure comprehensive scene understanding, we capture visual observations from two complementary perspective~\cite{aha}: a front view $v_1$ for tracking the robot embodiment's motion trajectory, and a top-down view $v_2$ for detecting perturbations $e_k$ during the skill execution $\ell_k$, as shown in \refFig{fig:vlm}. These dual views enable comprehensive workspace monitoring, supporting robust detection of perturbations and skill execution status.

\noindent\textbf{Query Structure.} To optimize computational cost while preserving essential workspace information, we subsample $N$ (default $N$=4) image frames from videos captured at 20 FPS for each view. 
Let $\mathbf{v}_{1,k}$ and $\mathbf{v}_{2,k}$ denote the sampled frames from the front and top-down views, respectively, during the skill execution of $\ell_k$. 
Each sequence includes the start and end frames, with intermediate frames randomly selected to provide contextual visual information.

Given the skill instruction $\ell_{k}$ from the planner $\mathcal{T}$, the actor executes the skills and captures visual observations $\mathbf{v}_{1,k}$ and $\mathbf{v}_{2,k}$ for visual queries. 
Upon execution, two targeted VQA language prompts ($Q_1$ and $Q_2$) are automatically generated based on $\ell_k$ to facilitate structured assessment and analysis.

\begin{itemize}[leftmargin=0.35cm]
    \item $Q_1$: `Did the robot successfully execute the action $\ell_{k}$?' Answer: `The action \{\textit{succeeded / failed}\}'. 
    
    \item $Q_2$: `Did any perturbation occur during the execution of $\ell_{k}$?' Answer: `No perturbation occurred' / `A never-seen \textit{[object]} appeared at \textit{[location]}' for ADD cases / similar descriptions generated for RMV and DIS perturbations.
\end{itemize}
Here, $Q_1$ evaluates the execution status of the skill instruction $\ell_{k}$, while $Q_2$ detects and characterizes perturbations, providing essential information to guide corrective re-planning if needed.

We formalize this global monitoring process as follows: 
\begin{equation} 
r_{1,k} = \text{VLM}(Q_1, \mathbf{v}_{1,k}), \quad r_{2,k} = \text{VLM}(Q_2, \mathbf{v}_{2,k}), \label{eq:vlm} 
\end{equation} 
where $r_{1,k}$ and $r_{2,k}$ represent the VLM's assessments of $\ell_k$'s execution results and detected perturbations $e_k$, respectively.

\noindent\textbf{Fine-Tuning VLM.} To facilitate robust TLP detection and execution assessment, we construct a diverse dataset of labeled episodes in a simulated environment~\cite{cliport}.
Each episode pairs execution outcomes with a single perturbation type ($\mathbf{E}$: ADD, RMV, or DIS) and includes dual-view observations along with structured textual descriptions across various task scenarios.
Leveraging this dataset, we fine-tune multiple open-source VLMs~\cite{openflamingo, blip_3, vila} using Supervised Fine-Tuning~\cite{vila, llava}, with a comparative evaluation presented in Sec.~\ref{sec:evaluation}.

\begin{figure}
    \centering
    \includegraphics[width=1\linewidth]{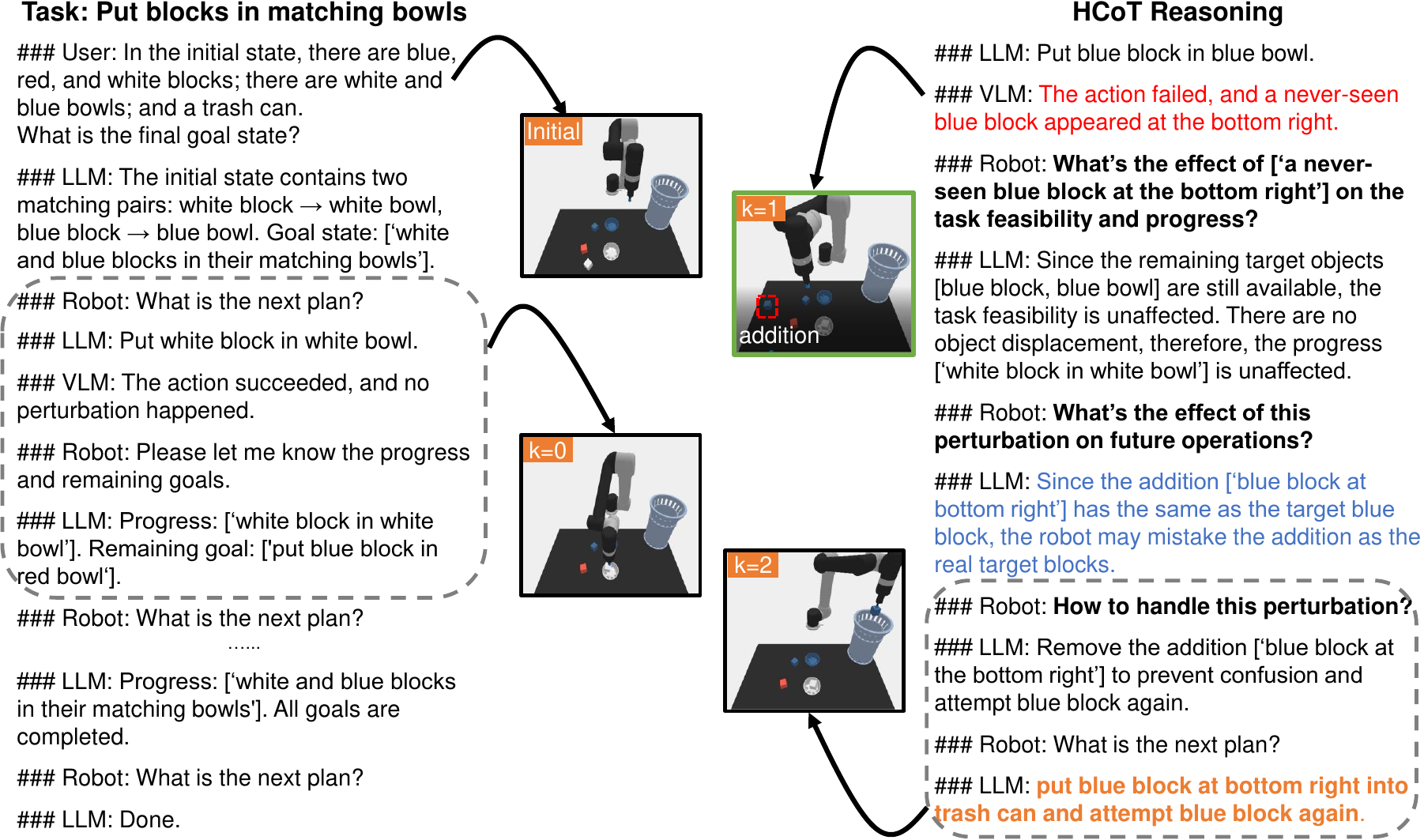}
    \vspace{-20pt}
    \caption{Prompt structure with HCoT reasoning. Illustrated with the matching task~\cite{ground_decoding}, our prompt consists of the following tags: ``User" to provide the initial state description, ``Robot" for fixed monologue instructions, and ``VLM" for updates from the VLM. Upon detecting a perturbation $e_k$, the LLM planner is prompted to analyze $e_k$ through multiple CoT steps, following a layered structure (feasibility → progress → operation) to generate the corrective plan.}
    \label{fig:prompt}
    \vspace{-15pt}
\end{figure}

\subsection{Textual Prompt for Corrective Re-planning of LLM}
To manage detected TLP and skill execution failures, we introduce a structured prompt template for the LLM-based planner $\mathcal{T}$ alongwith a Hierarchical Chain-of-Thought (HCoT) reasoning mechanism to facilitate corrective re-planning.

\noindent\textbf{In-context Learning for Task Planning.} Following prior studies~\cite{saycan, ground_decoding}, we employ instruction-tuned LLMs~\cite{llama, llava} as the high-level task planner $\mathcal{T}$, leveraging example episodes to guide in-context learning. 
Our prompt structure includes an initial scene description and step-by-step example episodes demonstrating task planning and expected outcomes.


At each planning step $k$, $\mathcal{T}$ generates the next skill instruction $\ell_k$ in response to the query, `What is your next plan?', informed by real-time feedback from the VLM. 
This feedback, consisting of updates $r_{1,k}$ and $r_{2,k}$ on execution outcomes and detected perturbations, enables $\mathcal{T}$ to evaluate task progress, adjust stratigies, and iteratively execute a closed-loop cycle of $\langle$execution, perception, reasoning$\rangle$~\cite{inner}, continuing until a “done” signal is issued, indicating task completion.


When the perturbation $e_k$ is detected via the VLM (Eq.~\ref{eq:vlm}), $\mathcal{T}$ engages in an HCoT-based reasoning process to systematically assess $e_k$’s impact and facilitate adaptive re-planning.

\noindent\textbf{HCoT Reasoning.} 
To systematically assess the varying impact of different perturbations, as mentioned in Sec.~\ref{sec:task}, we employ a hierarchical reasoning approach in the planner $\mathcal{T}$.

Our HCoT process follows a multi-round Q\&A dialogue (illustrated in \refFig{fig:prompt}), guiding the planner $\mathcal{T}$ through a structured, layered CoT reasoning~\cite{cot} process. 
Upon detecting a perturbation $e_k$ via the VLM, the planner $\mathcal{T}$ evaluates its impact across three key dimensions:
\begin{itemize}[leftmargin=0.37cm]
    \item \textbf{Layer-1\hspace{0.1cm}(feasibility)}: Checks if the task is still achievable. 
    \item \textbf{Layer-2\hspace{0.1cm}(progress)}: Assesses if task progress is preserved.
    \item \textbf{Layer-3\hspace{0.1cm}(future operations)}: Evaluates potential impacts on subsequent skill instructions. 
\end{itemize}

At each reasoning step, detailed explanations are required before drawing conclusions, ensuring a structured and interpretable decision-making process.
This HCoT mechanism equips $\mathcal{T}$ with a comprehensive understanding of workspace changes, enabling effective and corrective re-planning.


Each episode in our experiments is mostly designed to include a single TLP instance.
To further assess robustness and generalization to more complex scenarios, we evaluate LangPert in cases where multiple perturbations occur simultaneously at step $k$.
This HCoT reasoning enables $\mathcal{T}$ to effectively handle such combined cases, demonstrating its adaptability to previously unseen scenarios, as shown in Sec.~\ref{sec:evaluation}.



\subsection{Low-Level Visual Policies for Skill Execution}
To handle perturbations $e_k$, the low-level visual policy $\pi$ must precisely manipulate specific objects, even when they appear visually similar to task-relevant ones.
To achieve this, we improve CLIPort’s flexibility by incorporating relational and regional descriptors, which specify an object’s location relative to others (e.g., `in the red bowl') and its spatial location on the table (e.g., `at the top left'), respectively. 
These descriptors enable $\pi$ to differentiate visually identical objects, ensuring precise selection and manipulation.
Additionally, we introduce a trash can in the workspace, allowing the policy to discard interfering objects. 
With these modifications in the policy and environment, $\pi$ can execute detailed instructions, such as `put the red block in the red bowl into the trash can'.

    \section{Evaluation}
    \label{sec:evaluation}
    \begin{table*}[t]
\small
    \centering
    \caption{Success rate (\%) on ADD (task-related) and DIS perturbations, evaluated under conditions with and without skill execution failures (\textbf{w/o F.} and \textbf{w/ F.}) with a 0.2 failure probability. \textbf{None} indicates without task-level perturbations and skill failures. 
    }
    \vspace{-10pt}
    \begin{tabu}{m{1.2cm}<{\centering}|m{2cm}<{\centering}|m{1.5cm}<{\centering} m{1.5cm}<{\centering} m{1.9cm}<{\centering} m{1.5cm}<{\centering}m{2.4cm}<{\centering} m{2.4cm}<{\centering} }
           \tabucline[0.8pt]{-} 
        \multirow{2}{*}{Tasks}&   \multirow{2}{*}{Type}  & \multicolumn{6}{c
    }{Sucess Rate ($\uparrow$)}\\
    ~&~&SayCan~\cite{saycan}&SateInfer&IM-VLM~\cite{inner}&Ours&\textcolor{gray!80}{IM-ORACLE~\cite{doremi}}&\textcolor{gray!80}{Ours (ORACLE)}\\
           \tabucline[0.8pt]{-} 
    \multirow{5}{*}{\textit{Matching}}&
    None&\textbf{86.7}   & 79.5   & 84.2 & 83.5   & \textcolor{gray!80}{86.3   } & \textcolor{gray!80}{86.4   }\\
    ~&ADD (w/o F.)&64.7   & 76.5   & 76.7   & \textbf{78.7   }& \textcolor{gray!80}{79.7   } & \textcolor{gray!80}{81.6   }\\
    ~&ADD (w/ F.)&58.3   & 57.3   & 75.9   & \textbf{77.1   }& \textcolor{gray!80}{78.6   }& \textcolor{gray!80}{79.3   } \\
    ~&DIS (w/o F.)&52.4   &57.4   & 81.2   & \textbf{82.0   }& \textcolor{gray!80}{83.3   } & \textcolor{gray!80}{85.0   }\\
    ~&DIS (w/ F.)&43.1   & 48.7   & 70.9   & \textbf{75.4   }& \textcolor{gray!80}{76.3   }& \textcolor{gray!80}{80.3   } \\
    \tabucline[0.8pt]{-} 
    \multirow{5}{*}{\textit{Pack-B}}&None&89.6   & 72.5   & \textbf{91.3   }& 91.1   & \textcolor{gray!80}{92.7   } & \textcolor{gray!80}{92.6   }\\
    ~&ADD (w/o F.)&65.4   & 70.7   & 72.8   & \textbf{74.9   }& \textcolor{gray!80}{80.7   } & \textcolor{gray!80}{82.4   }\\
    ~&ADD (w/ F.)&57.0   & 62.1   & 72.0   & \textbf{74.1   }& \textcolor{gray!80}{78.1   }& \textcolor{gray!80}{79.5   } \\
    ~&DIS (w/o F.)&58.7   & 57.5   & 62.4   & \textbf{68.1   }& \textcolor{gray!80}{75.1   } & \textcolor{gray!80}{81.3   }\\
    ~&DIS (w/ F.)&47.0   & 48.1   & 56.7   & \textbf{63.4   }& \textcolor{gray!80}{73.7   }& \textcolor{gray!80}{80.6   }\\
    \tabucline[0.8pt]{-} 
    \multirow{5}{*}{\textit{Pack-G}}&None&81.4   & 78.4   & 80.7   & \textbf{81.2   }& \textcolor{gray!80}{84.1   } & \textcolor{gray!80}{83.4   }\\
    ~&ADD (w/o F.)&60.7   & 70.3   & 71.3   & \textbf{72.7   }& \textcolor{gray!80}{74.2  } & \textcolor{gray!80}{75.3   }\\
    ~&ADD (w/ F.)&55.3   & 60.1   & 68.1   & \textbf{69.5   }& \textcolor{gray!80}{73.4   }& \textcolor{gray!80}{74.7   } \\
    ~&DIS (w/o F.)&49.2   & 67.0   & 68.8   &\textbf{73.7   }& \textcolor{gray!80}{80.1   } & \textcolor{gray!80}{81.7   }\\
    ~&DIS (w/ F.)&41.3   & 54.1   & 65.4   & \textbf{68.7   }& \textcolor{gray!80}{77.0   }& \textcolor{gray!80}{78.4   }\\
    \tabucline[0.8pt]{-} 
    \multirow{5}{*}{\textit{Stacking}}&None&35.0   & 56.0   & 75.4   & \textbf{76.8   }& \textcolor{gray!80}{77.5   } & \textcolor{gray!80}{78.2   }\\
    ~&ADD (w/o F.)&17.0   & 38.7   & 67.5   & \textbf{71.3   }& \textcolor{gray!80}{73.0  } & \textcolor{gray!80}{76.1   }\\
    ~&ADD (w/ F.)&15.0   & 27.5   & 66.6   & \textbf{68.5   }& \textcolor{gray!80}{71.8   }& \textcolor{gray!80}{75.4   }\\
    ~&DIS (w/o F.)&12.0   & 32.1   & 55.8   & \textbf{64.6   }& \textcolor{gray!80}{67.6   } & \textcolor{gray!80}{77.8   }\\
    ~&DIS (w/ F.)&9.0   & 23.5   & 51.7   &\textbf{61.2   }& \textcolor{gray!80}{64.3   }& \textcolor{gray!80}{76.8   }\\
   \tabucline[0.8pt]{-} 
    \end{tabu}
    \label{tab:addd_dis_table}
    \vspace{-15pt}
\end{table*}

\subsection{Experimental Setup}
\noindent\textbf{Environments and Tasks.} We conduct experiments with a Universal Robot UR5e (6-DOF) with a suction gripper in the Ravens~\cite{ravens} environment. 
We also position two cameras in the scene to capture the front-view and top-down observations.  

Our evaluation comprises four tabletop rearrangement tasks:
\begin{itemize}[leftmargin=0.5cm]
\item \textbf{\textit{Matching}}~\cite{ground_decoding,saycan}: Match blocks to bowls by color.

\item \textbf{\textit{Packing}}: Pack selected objects into a brown box with two object-type variations: \textbf{\textit{Pack-B}}, using colored blocks, and \textbf{\textit{Pack-G}}, using Google Scanned objects~\cite{cliport}.

\item \textbf{\textit{Stacking}}: Construct a three-layer (3-2-1) block pyramid on a stand, adhering to color-specific constraints.
\end{itemize}

\noindent\textbf{Handling Perturbations and Execution Failures.} 
To assess the robustness of existing and our planners against Task-Level Perturbations (TLP), we introduce Addition (ADD), Removal (RMV), and Displacement (DIS) perturbations in each task episode. 
The occurrence step $k$ is randomly assigned during execution and is denoted as $e_k$ in each episode.

In ADD and RMV cases, $e_k$ introduces a change requiring the planner to classify the perturbed object as either a task-related one or a distractor. For ADD cases, new objects are added at collision-free locations (e.g., on the table, inside boxes, or on the stand). 
If the addition is classified as a task-related one, the planner $\mathcal{T}$ outputs a corrective instruction to discard it into the trash can; otherwise, it is ignored.
In RMV cases, the planner determines whether a removed object is task-related. If its absence affects task completion, an ``alert" is issued; otherwise, execution proceeds as usual.
For DIS cases, the planner generates appropriate, corrective skill instructions to restore task progress smoothly.

To simulate execution failures, we introduce a 0.2 probability for each skill instruction that the grasped object may drop, requiring repeated attempts until successful completion.

\noindent\textbf{Evaluation.} In addition to the Success Rate (SR), we evaluate: 
(1) \textbf{“Alert” accuracy} in RMV cases, assessing the planner's ability to correctly issue an “alert” for task-related object removals while ignoring distractor removals; and 
(2) \textbf{Average Steps to Completion (ASC)} in ADD distractor cases, measuring the number of execution steps taken by the actor model before the planner issues a “done” signal, serving as an efficiency metric for handling distractor additions.
All metrics are averaged over 100 episodes to ensure statistical reliability.

\noindent\textbf{Experimental Details and Baselines.} Following the description in Sec.~\ref{sec:method}, we evaluate OpenFlamingo~\cite{openflamingo}, VILA~\cite{vila}, and BLIP~\cite{blip_3} as VLM candidates. For the task planner, we employ the Llama series~\cite{llama}, incorporating one example episode per TLP scenario to facilitate in-context learning.

We validate the effectiveness of LangPert against several LLM baselines, emphasizing the contributions of the VLM monitor and hierarchical chain-of-thought (HCoT) reasoning: (1) \textbf{SayCan}~\cite{saycan}: This method pre-generates skill instructions before task execution (act as a lower bound), without scene feedback or re-planning mechanisms. (2) \textbf{Inner Monologue (IM)}~\cite{inner}: LangPert’s main distinction from IM lies in our HCoT design, which supports more comprehensive reasoning of task progress under TLP and enables more reliable corrective strategies. By contrast, IM outputs a corrective plan directly, without the experiencing hierarchical reasoning steps. We evaluate two IM variants: \textbf{IM-VLM}, which uses the same VLM monitor as LangPert for perturbation descriptions and failure detection, and \textbf{IM-ORACLE}, which assumes ground-truth monitoring feedback to establish an upper performance bound. (3) \textbf{StateInfer}: This baseline provides the LLM planner with ground-truth state descriptions at each step’s start and end frames, detailing object locations and spatial relationships—similar to scene graph~\cite{reflect,sayplan} and multi-modal language model (MLM) methods~\cite{GPT_4V,manipulate_anything} for task state encoding. In contrast, LangPert uses a dual VLM to monitor real-time workspace dynamics, explicitly identifying both execution outcomes and TLP. To ensure a fair comparison, we provide StateInfer with additional episode examples for prompting, allowing the LLM to better interpret state information and infer outcomes for each TLP scenario through in-context learning. (4) \textbf{Ours (ORACLE)}: Our method replaces the VLM with ground-truth feedback, enabling a direct comparison with IM-ORACLE to test the effectiveness of our prompt design.


\begin{figure*}
    \centering
    \includegraphics[width=1\linewidth]
    {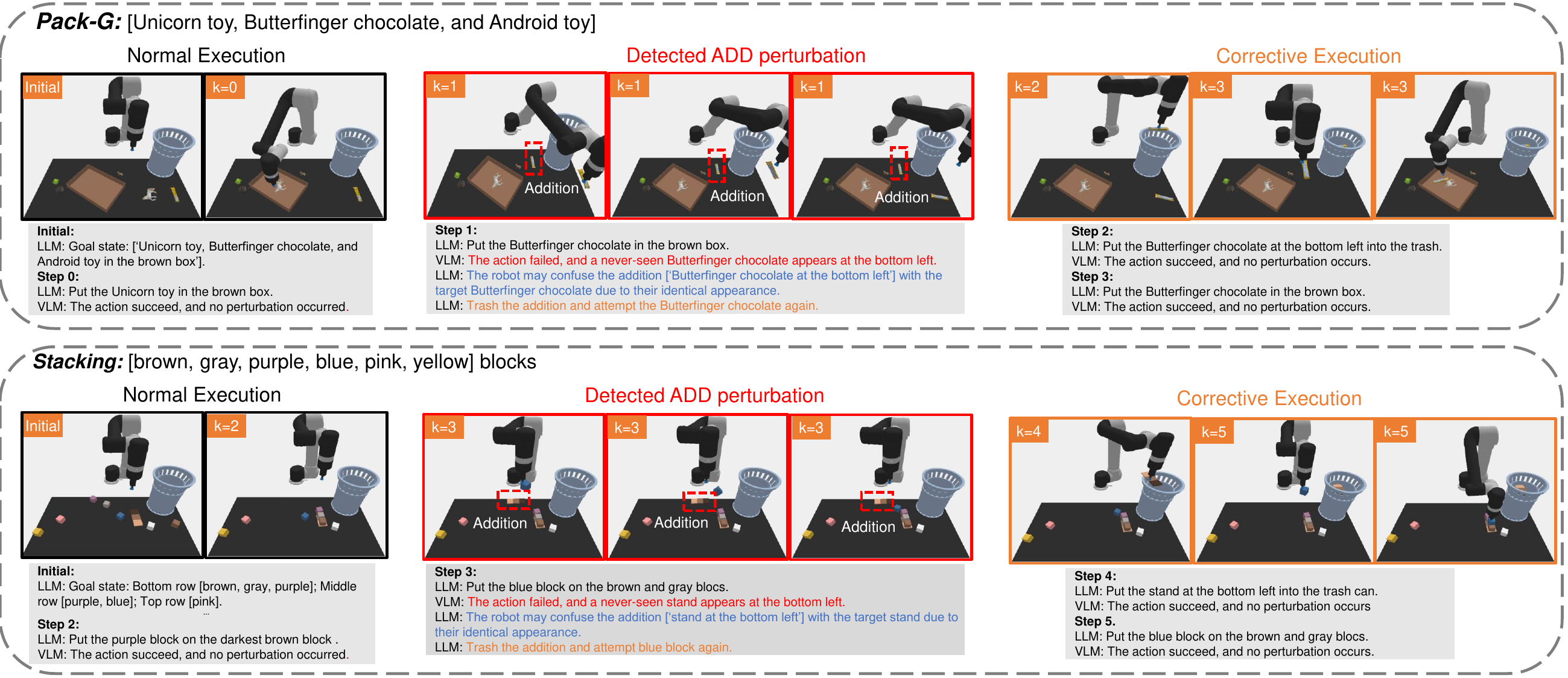}
    \vspace{-20pt}
    \caption{Qualitative Results in task-related ADD Scenarios. In the \textit{Pack-G} and \textit{Stacking} tasks (brackets are specified goal objects), the robot encounters concurrent ADD perturbations and skill execution failures at steps $k=1$ and $k=3$, where new objects (a Butterfinger chocolate and a stand, respectively) appear at the bottom left of the table (highlighted in red). Guided by VLM feedback, LangPert identifies these additions as potential interferences due to their similarity to goal objects and generates corrective strategies to instruct the robot to remove them before re-attempting the failed skill.}
    \label{fig:qalitative_result_1}
    \vspace{-15pt}
\end{figure*}
\subsection{Result Analysis}
After evaluating our LangPert framework across tasks and various TLP cases, we have the following observations:
\label{sec:results}

\noindent\textbf{\textit{1. LangPert can detect various TLP and generate adaptive, corrective plans.}} \refTab{tab:addd_dis_table} compares our method and baselines under None cases, which is in a static situation without any TLP and skill execution failure, as well as ADD (task-related) and DIS cases. 
LangPert demonstrates robust performance across all tasks, leveraging its VLM-based detection of task-level perturbations and skill-execution failure states.

In the None cases, LangPert performs similarly to IM-VLM while outperforming SayCan and StateInfer, confirming that it does not significantly alter execution in static, stable environments and primarily responds to perturbations when they occur. SayCan suffers a dramatic SR drop when perturbations and execution failures are introduced due to its lack of scene feedback and re-planning mechanisms, highlighting the necessity of these capabilities in dynamic environments.
In the ADD cases, LangPert performs comparably to Inner Monologue (IM), as indicated by results from IM-ORACLE vs. Ours (ORACLE) and IM-VLM vs. Ours. 
In the more complex DIS cases (multiple object may involved) requiring precise corrective instruction sequences, LangPert outperforms IM. Notable improvements include a nearly 10\% SR $\uparrow$ for the \textit{Stacking} task (51.7\% $\rightarrow$ 61.2\%), and approximately a 5\% $\uparrow$ in the \textit{Matching} (70.9\% $\rightarrow$ 75.4\%) and \textit{Pack-B} (56.7\% $\rightarrow$ 63.4\%) tasks. These results demonstrate that LangPert’s HCoT mechanism enhances the LLM planner’s ability to analyze perturbed task progress through our layered Q\&A process, resulting in more accurate multi-step corrective plans.
Please refer to a supplementary video for additional task demonstrations and intuitive understanding of perturbation handling processes.

\noindent\textbf{\textit{2. An independent perception model is necessary.}} While StateInfer (which uses a single LLM for both task planning and monitoring) demonstrates notable improvements (in \refTab{tab:addd_dis_table}, across tasks) over SayCan, its performance degrades more significantly in complex scenarios, such as DIS (w/ F.) cases, where skill execution failures and TLP co-occur. This suggests that a single LLM struggles to distinguish between overlapping task state changes caused by skill failures and perturbations, leading to reasoning confusion. An alternative approach~\cite{rt2,sc_mllm} is to fine-tune the LLM with extensive task state data from various perturbation scenarios, though this requires substantial data and high computational costs.

On the other hand, LangPert separates perception and planning and adopts independent language models for each functionality~\cite{doremi}. This framework allows the LLM planner to receive conclusive information from the VLM monitor, avoiding the need for complex perception-based reasoning. Besides, our VLM monitor distinguishes between execution failures and TLP by leveraging dual (front \& top-down) views to improve situational awareness. This structured separation reduces ambiguity between different types of state changes, simplifies VLM reasoning, and enhances perception robustness in handling cases where multiple factors occur simultaneously.

\begin{table}[b]
    \centering
    \caption{Success Rate (SR) and Average Steps to Completion (ASC) per episode when encountering distractor ADD perturbations (with 0.2 skill execution failure probability).}
    \vspace{-5pt}
    \begin{tabular}{c|cc|cc}
        \hline
        \multirow{2}{*}{Task} &\multicolumn{2}{c|}{IM-VLM~\cite{inner}}  & \multicolumn{2}{c}{Ours} \\
        ~&SR ($\uparrow$) & ASC ($\downarrow$) & SR ($\uparrow$)& ASC ($\downarrow$)\\
        \hline
        \textit{Matching}&78.4   &5.1   &80.9   &\textbf{4.4   }\\
        \textit{Pack-B}&79.4   &9.1   &86.5   &\textbf{8.5   }\\
        \textit{Pack-G}&73.3   &5.0   &76.8   &\textbf{4.4   }\\
        \textit{Stacking}&70.1   &7.9   &74.3   &\textbf{7.3   }\\
        \hline
    \end{tabular}
    \label{tab:distractor_add}
    \vspace{-13pt}
\end{table}

\begin{table}[b]
\small
    \centering
    \caption{``alert" accuracy analysis under RMV perturbations. In this table, `task-related' cases are considered \textit{positives}, while `distractor' cases are treated as \textit{negatives}.}
    \vspace{-5pt}
    \begin{tabular}{c|cccc|cccc}
        \hline
         \multirow{2}{*}{Task}& \multicolumn{4}{c}{IM-VLM} &\multicolumn{4}{c}{Ours}\\
       ~&TP&FN&FP&TN&TP&FN&FP&TN\\
         \hline
        \textit{Matching}&67 &33&61&39&92 &8&12&88\\
        \textit{Pack-B}&72 &28&78&22&88 &12&14&86\\
        \textit{Pack-G}&73 &27&71&29&97 &3&6&94\\
        \textit{Stacking}&47&53&51&49&87 &13&17&83\\
         \hline
    \end{tabular}
    \vspace{-13pt}
    \label{tab:alert}
\end{table}

\begin{table}[t]
\small
    \centering
    \caption{Success rate ($\uparrow$) in mixed (ADD \& DIS) perturbation scenarios.}
    \vspace{-5pt}
    \begin{tabular}{c|cccc}
        \hline
         Method& \textit{Matching} &\textit{Pack-B}&\textit{Pack-G}&\textit{Stacking}\\
         \hline
         IM-ORACLE&65.5&62.5&66.1&60.7\\
        Ours (ORACLE)&\textbf{75.4}&\textbf{70.4}&\textbf{72.1}&\textbf{68.9}\\
         \hline
    \end{tabular}
    \vspace{-10pt}
    \label{tab:unseen}
\end{table}

\begin{figure}[!t]
    \centering
    
    \includegraphics[width=1\linewidth]{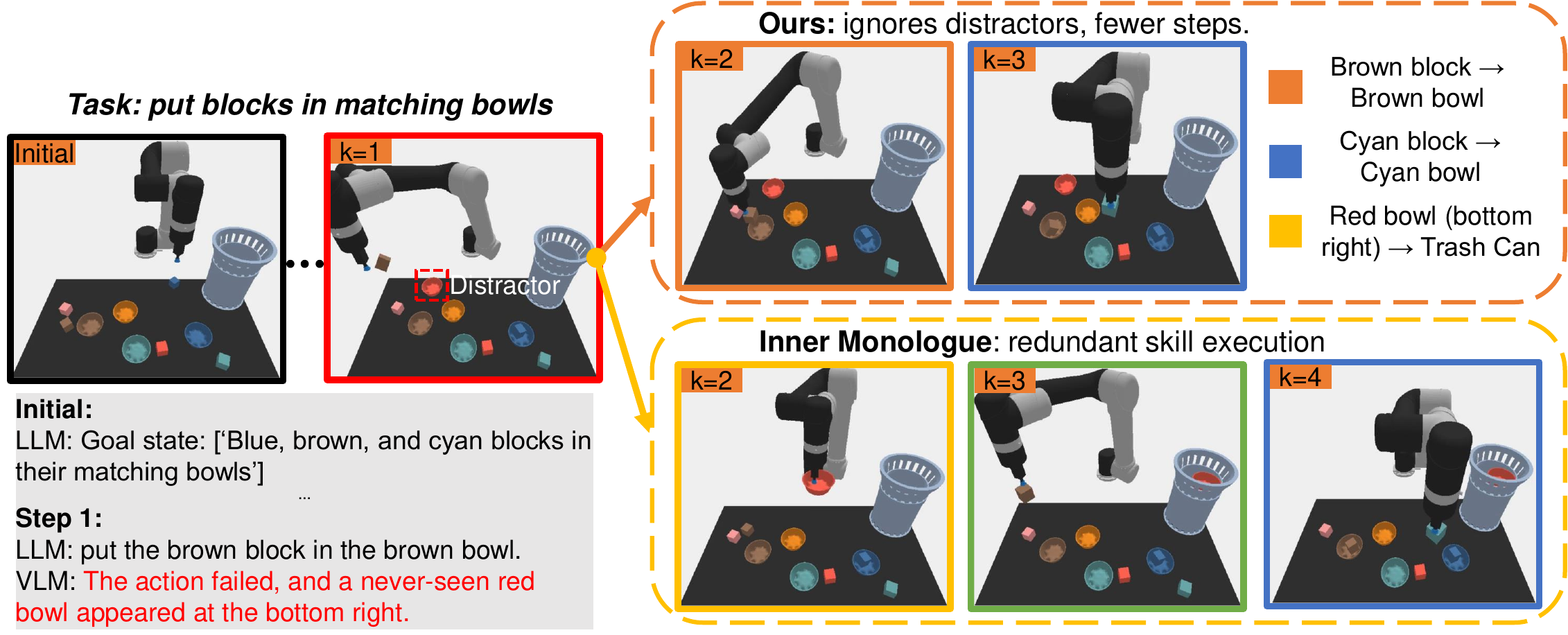}
    \vspace{-15pt}
    \caption{Comparison in an ADD distractor case. Step 1 (k=1) shows a failed execution alongside the appearance of a distractor (red bowl, marked with a red box). LangPert correctly identifies the distractor as task-irrelevant and proceeds without interruption, whereas Inner Monologue~\cite{inner} introduces an unnecessary instruction based on the same VLM feedback.}
    \label{fig:contextual_understanding}
        \vspace{-10pt}
\end{figure}

\noindent\textbf{\textit{3. HCoT enhances contextual understanding.}}  We evaluate the contextual understanding capabilities of LangPert (ours) and IM by examining their corrective planning performance across different perturbation scenarios, as follows:

\begin{itemize}[leftmargin=0.5cm]
    \item \textbf{Improved efficiency in ADD distractor cases.} \refTab{tab:distractor_add} demonstrates that LangPert achieves higher SR and lower ASC (fewer execution steps) compared to IM-VLM in ADD distractor cases, where newly introduced objects are task-irrelevant and do not affect the low-level policy's execution. A qualitative example is illustrated in~\refFig{fig:contextual_understanding}.
    
    \item \textbf{Accurate RMV classification.} \refTab{tab:alert} compares our method with IM in handling RMV perturbations, treating the removal of task-relevant objects as positives and distractors as negatives. LangPert demonstrates higher precision and recall in generating ``alert" signals.

    \item \textbf{Generalization in multiple, mixed TLP cases.} We evaluate model performance in mixed scenarios where ADD (task-related) and DIS occur simultaneously at step $k$ (see \refTab{tab:unseen}). Ours achieves higher SR than IM across tasks, demonstrating enhanced reasoning capabilities enabled by HCoT. Qualitative results reveal that the IM model, without HCoT, typically addresses only one perturbation in combined cases, resulting in inferior performance. Please refer to the supplementary video for intuitive understanding. 

\end{itemize}
Overall, these results validate the effectiveness of our HCoT design. Through layered analysis (feasibility → progress → future operation), LangPert accurately discerns the impacts of different perturbations, reducing redundant skill executions and improving task efficiency and generalization, even under novel conditions. The step-by-step CoT reasoning structure enables a deeper understanding of task state changes, empowering the planner to make more reliable corrective decisions.

\begin{table}[b]
\small
    \centering
    \caption{Effects of the VLM monitor (on \textit{Stacking} task).}
    \vspace{-5pt}
    \begin{tabular}{c|c c c}
        \hline
         \multirow{2}{*}{Type}& \multicolumn{3}{c}{Sucess Rate ($\uparrow$)} \\
       ~&OpenFlamigo~\cite{openflamingo}&VILA~\cite{vila}&BLIP-3~\cite{blip_3}\\
         \hline
        None&74.5 &\textbf{77.0}&76.8\\
        ADD (w/o F.)&60.1 &68.7&\textbf{71.3}\\
        ADD (w/ F.)&57.8 &64.6&\textbf{68.5}\\
        DIS (w/o F.)&59.4 &62.3&\textbf{64.6}\\
        DIS (w/ F.)&52.1 &60.9&\textbf{61.2}\\
         \hline
    \end{tabular}
    \label{tab:abl_vlm}
    \vspace{-13pt}
\end{table}

\begin{table}[b]
\small
    \centering
    \caption{Effects of Llama~\cite{llama} planners (on \textit{Stacking} task).}
    \vspace{-5pt}
    \begin{tabular}{c|ccc}
         \hline
         \multirow{2}{*}{Type}& \multicolumn{3}{c}{Sucess Rate ($\uparrow$)} \\
       ~&LLama 2-7B&LLama 3-8B&LLama 3.1-8B\\
         \hline
        None&72.7 &\textbf{76.8}&\textbf{76.8}\\
        ADD (w/o F.)&\textbf{71.5}&\textbf{71.5}&71.3\\
        ADD (w/ F.)&66.4&67.7&\textbf{78.5}\\
        DIS (w/o F.)&58.3 &63.8&\textbf{64.6}\\
        DIS (w/ F.)&49.7 &55.8&\textbf{61.2}\\
         \hline
    \end{tabular}
    \label{tab:abl_planner}
        \vspace{-13pt}
\end{table}

\subsection{Ablation Studies}
Lastly, we evaluate a variety of Vision Language Models (VLMs) and Llama planners to analyze their capabilities in detecting perturbations and generating corrective plans.

\noindent\textbf{Effects of VLMs.} We compare three candidate models: OpenFlamingo-3B~\cite{openflamingo} (OpenFlamingo), VILA-1.5-3B~\cite{vila} (VILA), and BLIP-3-4B~\cite{blip_3} (BLIP-3). The evaluation results are presented in \refTab{tab:abl_vlm}. BLIP-3 consistently outperforms the others in identifying and describing ADD and DIS, as well as in detecting execution failures. The superior performance of BLIP-3 can be attributed to its larger model size and scalable encoding strategy~\cite{any_resolution}, which effectively captures detailed semantic and spatial information from sequential image frames.

\noindent\textbf{Effects of LLM Planner.} We also compare different variants of the Llama planner, as shown in \refTab{tab:abl_planner}. Llama 3.1 demonstrates the best overall performance. While Llama 2 and Llama 3 exhibit comparable planning capabilities under the None cases (i.e., no TLP and flawless skill execution), the advanced textual reasoning capabilities of Llama 3.1 enable it to analyze and respond more effectively to complex perturbation scenarios. This advantage is particularly pronounced in DIS (w/ F.) cases, where the planner must generate coherent, sequential corrective instructions.



\noindent\textbf{Limitations \& Future Work.} 
While LangPert has demonstrated effectiveness in addressing TLP across various rearrangement tasks, several limitations remain. First, the VLM displays inferior results in unseen perturbations that require fine-grained contextual understanding, e.g., precise object counting and spatial reasoning. This challenge is pronounced in cases involving multiple object additions and complex multi-layer stacking tasks. Besides, our experiments primarily focus on pick-and-place primitives, which may constrain the diversity of manipulation tasks. To overcome this limitation, we plan to expand the low-level policy’s skill repository by integrating complementary policy models~\cite{peract}, enabling more complex and adaptive real-world applications. Furthermore, future work will incorporate additional sensing modalities, such as height maps~\cite{ravens}, to enhance the VLM’s reasoning capabilities and improve its generalization to unseen scenarios.



    \section{Conclusion}
    \label{sec:conclusion}
    We introduced LangPert, a language-based framework designed to address Task-Level Perturbations (TLP)—including unexpected object additions, removals, and displacements—in tabletop rearrangement tasks. LangPert leverages a Vision Language Model (VLM) for real-time execution monitoring and TLP detection, while incorporating a Hierarchical Chain-of-Thought (HCoT) reasoning mechanism to enhance the Large Language Model (LLM) planner’s contextual understanding and adaptive, corrective plan generation. Experimental results demonstrate that LangPert effectively handles diverse TLP scenarios across tasks, with promising generalization to previously unseen conditions. Future work will focus on real-world deployment and further improvements to VLM capabilities.

    {
        \bibliographystyle{IEEEtranS}
        \bibliography{egbib}
    }
\end{document}